\documentclass[runningheads]{llncs}

\usepackage{eccv}

\usepackage{eccvabbrv}

\usepackage{graphicx}
\usepackage{booktabs}
\usepackage{multirow}
\usepackage{float}

\usepackage[accsupp]{axessibility}  

\usepackage{hyperref}

\usepackage{orcidlink}

\begin{document}

\title{Gaussian Heritage: 
3D Digitization of Cultural Heritage with Integrated Object Segmentation \thanks{This project has received funding from the European Union's Horizon research and innovation programme under grant agreement No 101079116 and No 101079995.}}

\titlerunning{Gaussian Heritage}

\author{Mahtab Dahaghin\orcidlink{0009-0001-5559-4884} \and
Myrna Castillo \orcidlink{0000-0001-6814-261X} \and
Kourosh Riahidehkordi\orcidlink{0009-0003-2780-9942} \and
Matteo Toso\orcidlink{0000-0002-8990-7156} \and
Alessio Del Bue\orcidlink{0000-0002-2262-4872}}

\authorrunning{M.~Dahaghin~\etal}

\institute{Pattern Analysis and Computer Vision (PAVIS)\\Istituto Italiano di Tecnologia (IIT),
Genoa, Italy\\
\tt\small \{name.surname\}@iit.it\\
}

\maketitle

\begin{abstract}
  The creation of digital replicas of physical objects has valuable applications for the preservation and dissemination of tangible cultural heritage. However, existing methods are often slow, expensive, and require expert knowledge. We propose a pipeline to generate a 3D replica of a scene using only RGB images (\eg photos of a museum) and then extract a model for each item of interest (\eg pieces in the exhibit). We do this by leveraging the advancements in novel view synthesis and Gaussian Splatting, modified to enable efficient 3D segmentation. This approach does not need manual annotation, and the visual inputs can be captured using a standard smartphone, making it both affordable and easy to deploy. We provide an overview of the method and baseline evaluation of the accuracy of object segmentation. The code is available at \url{https://mahtaabdn.github.io/gaussian_heritage.github.io/}. 
\end{abstract}

\section{Introduction}
\label{sec:intro}
\begin{figure}[tb!]
    \centering
    \includegraphics[width=.85\textwidth] {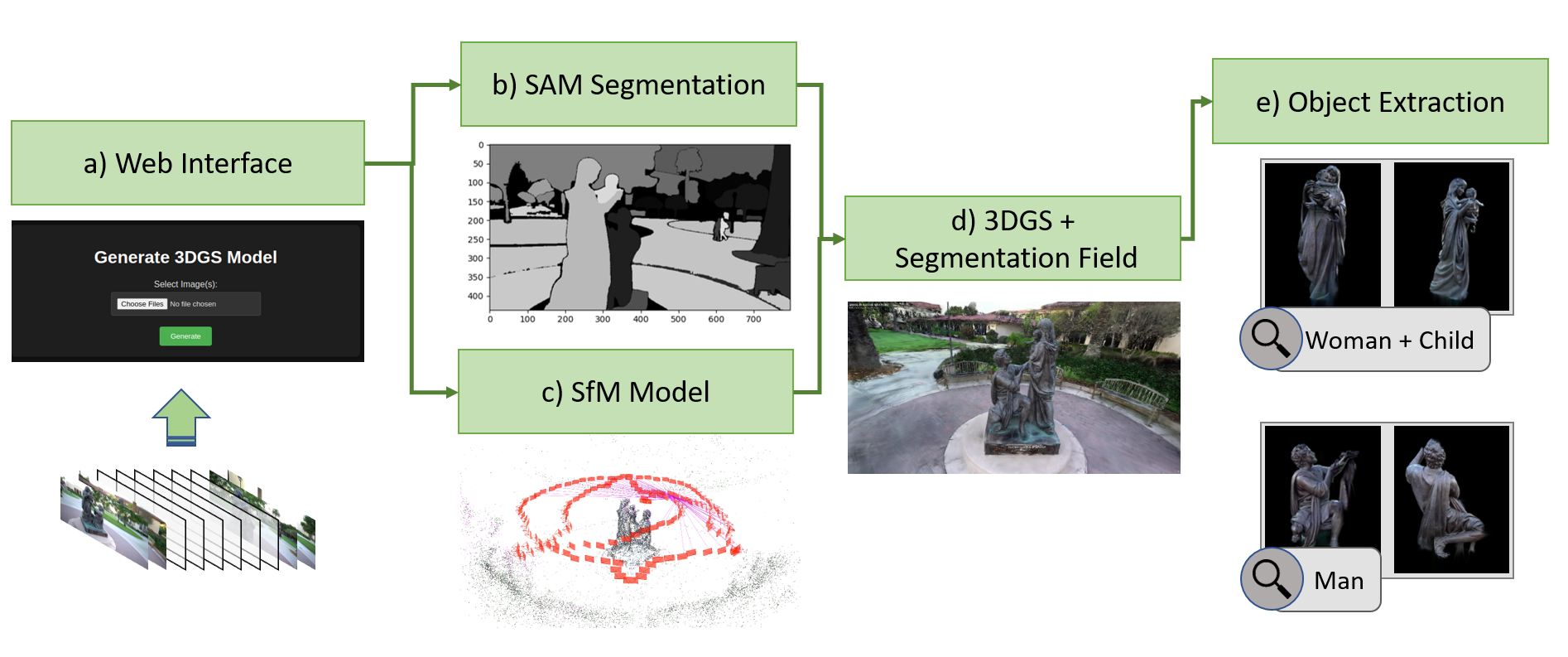}
    \caption{Given a set of images, we a) use a web interface to upload them to a local server, where they are processed to generate b) 2D instance segmentation masks and c) a sparse 3D model. Using these, we train d) a model that captures appearance and 3D segmentation of the scene, from which we can e) extract a model for each object. }
    \label{fig:themethod}
\end{figure}

Computer Vision has many interesting applications in the diffusion and preservation of Cultural Heritage (CH), including the digitization of art. Creating virtual replicas enables designing virtual museum tours~\cite{louvre}; creating catalogs for archival purposes~\cite{noho}; or aiding art restoration by reassembling digital copies before handling the potentially delicate remains~\cite{repair}. 
However, current art digitization methods are slow, expensive, and often rely on expert technicians and specialized equipment~\cite{versus, cultarm}, limiting the widespread adoption of these techniques. To address these limitations, we propose a simpler framework, as shown in Fig.~\ref{fig:themethod}. Using only RGB images captured by a standard smartphone, we leverage advancements in novel view synthesis~\cite{mildenhall2020nerf,kerbl3Dgaussians, gaussian_grouping,silva2024contrastivegaussianclusteringweakly} to generate a photo-realistic model of the scene that also provides segmentation of the CH items of interest. 
We then use open-vocabulary object detector Grounding DINO~\cite{liu2023grounding}  to identify CH items in arbitrary views and extract the corresponding components from the model. This approach is a significant step towards making CH digitization also accessible to institutions with limited resources. 

The main contributions of our work are:
    \emph{i)} a pipeline to generate instance-aware 3D Gaussian Splatting models, enabling segmentation of the model into its components;
    and \emph{ii)} an automated reconstruction and segmentation process, packaged in an easily deployable Docker container.


\section{Related Work}
\label{sec:relatedwork}

Classical approaches model 3D scenes as point-clouds~\cite{schoenberger2016sfm} or 3D meshes~\cite{Wang_2018_ECCV}, but recent models like NeRF~\cite{mildenhall2020nerf} use a continuous volumetric function that maps position and viewing direction to density and color, learned from a limited number of images with known poses. This enables the generation of photo-realistic images from any point of view. 
Alternatively, 3D Gaussian Splatting (3DGS)~\cite{kerbl3Dgaussians}, replaces the continuous representation with a discrete number of 3D Gaussians with different sizes, orientations, and view-dependent colors, achieving both competitive training times and real-time rendering at higher image resolution.

These models can be combined with the advancements in foundation models~\cite{li2022languagedriven,caron2021emerging,kirillov2023segany}, that give access to accurate, open-vocabulary segmentation masks, to embed the segmentation masks generated by LLMs into NeRF~\cite{lerf2023,Zhi:etal:ICCV2021} and 3DGS models~\cite{qin2023langsplat,cen2023segment}. This is done using well-established strategies to embed 2D masks into 3D models,  \eg, by ``lifting'' 2D masks by back-projection~\cite{Genova2021Learning3S} or using contrastive learning between 2D and 3D features~\cite{Liu2021ContrastiveMF,Sautier2022ImagetoLidarSD}. Building on these works, we automatically generate SAM masks for each input image, using a contrastive loss to embed such information into a 3DGS model. Unlike existing methods, however, we complement the contrastive loss with other forms of supervision, such as regularization based on the spatial distance between the Gaussians; and instead of learning the visual and segmentation information separately, we jointly optimize all information stored in the 3D Gaussians.

\section{Methodology}
\label{sec:methodology}


Our pipeline for scene modeling and segmentation is composed of three key stages: \emph{i)} training the model, \emph{ii)} rendering 2D masks, and \emph{iii)} 3D extraction.

\paragraph{Model Training and Optimization.} Our method represents the scenes as clouds of 3D Gaussians that model geometry, appearance, and instance segmentation, using a modified 3DGS~\cite{kerbl3Dgaussians} model. Each Gaussian is augmented with a 16-dimensional feature vector ($\mathbb{R}^{16}$) to encode segmentation information, extrapolated from 2D segmentation masks generated by ViT-H SAM~\cite{kirillov2023segany}. We simultaneously optimize the segmentation feature vector and the standard 3DGS parameters using a combination of three losses:


\begin{enumerate}
    \item The standard 3DGS loss to optimize geometry and appearance ($\mathcal{L}_{rendering}$).
    \item A contrastive clustering loss~\cite{ying2023omniseg3d} to align rendered features with 2D segmentation masks 
    ($\mathcal{L}_{CC}$).
    \item A spatial-similarity regularization to enforce spatial continuity of feature vectors, encouraging adjacent 3D Gaussians to have consistent segmentation features while discouraging distant ones from having similar features ($\mathcal{L}_{reg}$).
\end{enumerate}

The total loss is composed of  rendering, clustering, and regularization terms:

\begin{equation}
     \mathcal{L} = \mathcal{L}_{rendering} + \lambda_{clustering}\mathcal{L}_{CC} + \mathcal{L}_{reg},
 \end{equation}

\paragraph{2D Mask Rendering.} The trained model can be used to render both semantic and instance 2D segmentation masks. To obtain \emph{instance segmentation}, we select an arbitrary viewpoint and splat on it the segmentation features. This results in a feature vector for each pixel, and we can assign to the same instance all pixels whose feature vectors have a cosine similarity below a given threshold $t$, which we empirically set at \( t = 0.7 \).
The \emph{semantic segmentation} is instead obtained using text prompts. Given \emph{a)} a text prompt related to an object in the scene, we \emph{b)} randomly sample various viewpoints and select in each a bounding box using Grounding DINO~\cite{liu2023grounding}. Then \emph{c)} we use SAM to refine the bounding boxes into 2D masks and \emph{d)} select the best fitting instance segmentation mask.

\paragraph{3D Extraction.} To achieve 3D segmentation, we render an initial view and use GroundingDINO and SAM to generate a mask for the target object based on a user-provided text prompt. We extract a feature prompt from this mask and perform similarity-based segmentation of 3D Gaussians using cosine similarity. To improve coherence, we apply a convex hull algorithm~\cite{gaussian_grouping} on the masked points. We then modify the original model by extracting the subset of Gaussians corresponding to the refined 3D mask. Finally, we save the modified Gaussian model and render the scene from multiple viewpoints. 
ambra marascio

\section{Early Results}
\label{sec:exp:quantitative}

\begin{table}[ht]
\scriptsize
\centering
    \caption{Comparison of semantic segmentation on the LERF-Mask and 3D-OVS datasets. We report average mIoU and mBIoU for each dataset (higher is better). 
    }

\begin{tabular}{|c|c|c|c|c|c|}
\hline
Dataset & Metric  & LeRF~\cite{lerf2023} & Gaussian Grouping~\cite{gaussian_grouping} & LangSplat~\cite{qin2023langsplat} & Ours \\ 
\hline
\multicolumn{1}{|c|}{\multirow{2}{*}{LeRF-Mask}} & mIoU  &  37.2    &         72.8          &      44.5     &   \textbf{80.3}   \\ 
\cline{2-6}
\multicolumn{1}{|c|}{}                                     & mBIoU &   29.3   &         67.6          &      39.8     &  \textbf{76.9} \\ 
\hline
\multicolumn{1}{|c|}{\multirow{2}{*}{3D-OVS}}    & mIoU  &   54.8   &          82.9         &      67.8     &   \textbf{87.5} \\ 
\cline{2-6}
\multicolumn{1}{|c|}{}                                     & mBIoU &   n.a.   &            78.4       &      59.6     &   \textbf{81.1} \\ 
\hline

\end{tabular}%
\label{tab:iou}

\end{table}

\begin{figure}[tb!]
      \centering
      \includegraphics[width=0.8\textwidth] {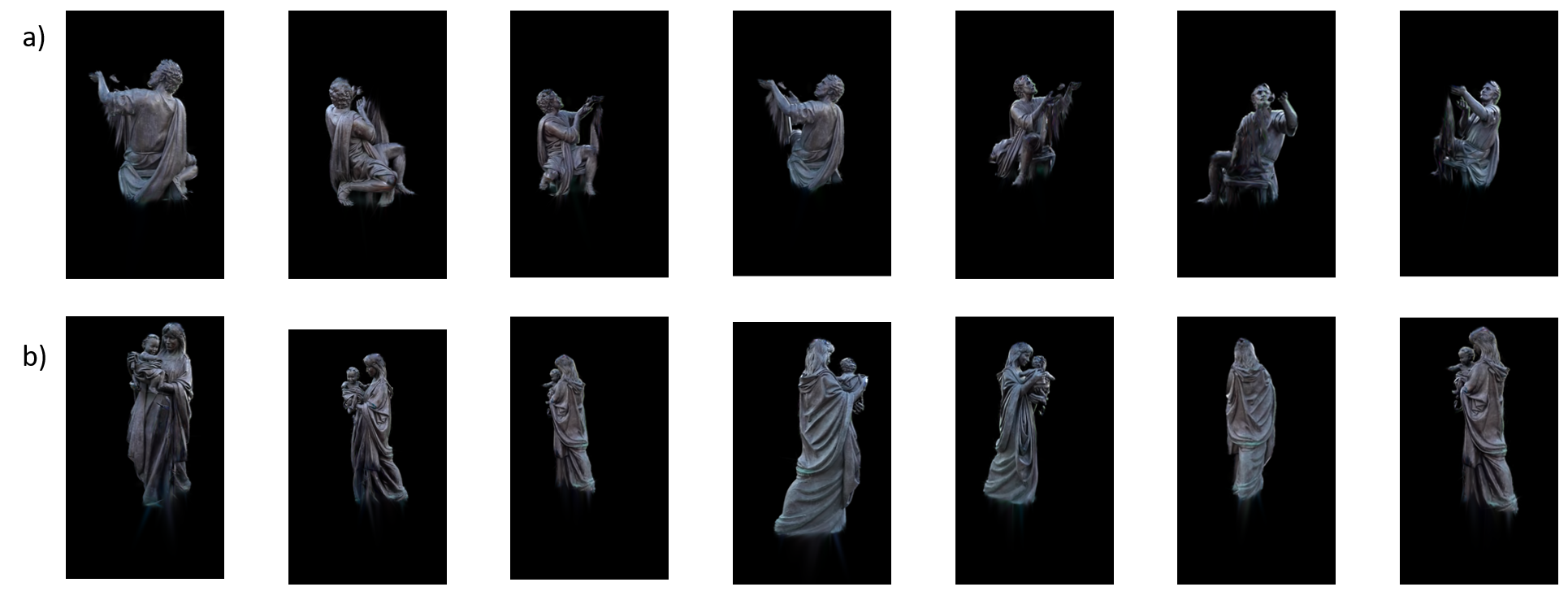}
      \caption{Sample view extracted from the ``Family'' scene of the Tanks and Temples dataset~\cite{Knapitsch2017}, using the labels a) ``man statue'' and b) ``mother and baby statue''.
     }
     \label{fig:maryonacross}
 \end{figure}

To support the proposed pipeline and methodology, we first explore the accuracy of the segmentation field learned by our method. We provide an evaluation benchmark by testing the method on two datasets commonly used in 3D scene segmentation with 3DGS: LERF-Mask~\cite{lerf2023} and 3D-OVS~\cite{liu2023weakly}. We compare our results against the segmentation accuracy of LeRF~\cite{lerf2023}, Gaussian Grouping~\cite{gaussian_grouping} and LangSplat~\cite{qin2023langsplat}. In Table~\ref{tab:iou}, we report the average performance in terms of mean intersection over union (mIoU) and mean boundary intersection over union (mBIoU), showing how our model provides more accurate segmentations.
Figure~\ref{fig:maryonacross} then provides qualitative results to showcase how our method can extract accurate models for individual objects in the scene. 
Using as input 150 images from the ``Family'' scene of the Tanks and Temples dataset~\cite{Knapitsch2017}, we train a model of the full scene and extract the Gaussians corresponding to two text prompts. 

\section{Conclusion}
\label{sec:exp:qualitative}
We have deployed a pipeline that can extract accurate 3D models of any object in a scene using only RGB images as input. Future work will focus on evaluating the model in 3D segmentation tasks, on-site testing in museums, and other possible target beneficiaries of this approach. The code for the pipeline is available at \url{https://mahtaabdn.github.io/gaussian_heritage.github.io/}. 

\bibliographystyle{splncs04}
\bibliography{main}
\end{document}